# Infinite Hidden Relational Models

**Zhao Xu**
Institute for Computer Science
University of Munich
Munich, Germany

**Volker Tresp, Kai Yu**
Information and Communications
Corporate Technology, Siemens AG
Munich, Germany

**Hans-Peter Kriegel**
Institute for Computer Science
University of Munich
Munich, Germany

## Abstract

Relational learning analyzes the probabilistic constraints between the attributes of entities and relationships. We extend the expressiveness of relational models by introducing for each entity (or object) an infinite-dimensional latent variable as part of a Dirichlet process (DP) mixture model. We discuss inference in the model, which is based on a DP Gibbs sampler, i.e., the Chinese restaurant process. We extended the Chinese restaurant process to be applicable to relational modeling. We discuss how information is propagated in the network of latent variables, reducing the necessity for extensive structural learning. In the context of a recommendation engine our approach realizes a principled solution for recommendations based on features of items, features of users and relational information. Our approach is evaluated in three applications: a recommendation system based on the MovieLens data set, the prediction of gene function using relational information and a medical recommendation system.

## 1 Introduction

Relational learning (Dzeroski & Lavrac, 2001; Raedt & Kersting, 2003; Wrobel, 2001; Friedman et al., 1999) is an object oriented approach that clearly distinguishes between entities (e.g, objects), relationships and their respective attributes and represents an area of growing interest in machine learning. A simple example of a relational system is a recommendation system: based on the attributes of two entities, i.e. of the user and the item, one wants to predict relational attributes like the preference (rating, willingness to purchase, ...) of this user for this item. In many circumstances, the attributes of the entities are rather weak predictors in which case one can exploit the known *relationship* attributes to predict unknown entity or relationship attributes (Yu et al., 2004). In recommendation systems, the latter situation is often referred to as collaborative filtering. Although the unique identifier of an entity to which a relationship exists might often be used as a feature, it has the disadvantage that it does not permit the generalization to new entities. From this point of view it is more advantageous to introduce a latent variable representing unknown attributes of the entities, which is the approach pursued in this paper. Attributes of entities are now children of the corresponding entity latent variable and attributes of relationships are children of the latent variables of the entities participating in the relationship. By introducing the latent variables the ground network forms a relational network of latent variables. Thus, our *hidden relational model* can be viewed on as a direct generalization of hidden Markov models used in speech or hidden Markov random fields used in vision (such models are discussed, for example, in Yedidia et al. 2005). As in those models, information can propagate across the network of latent variables in the hidden relational model, which reduces the need to extensive structural model selection. Structural model selection is a major problem in relational learning due to the exponentially many features an attribute might depend on. Thus information about my grandfather can propagate to me via the latent variable of my father.

Since each entity class might have a different number of states in its latent variables, it is natural to allow the model to determine the appropriate number of latent states in a self-organized way. This is possible by embedding the model in Dirichlet process (DP) mixture models, which can be interpreted as a mixture models with an infinite number of mixture components but where the model, based on the data, automatically reduces the complexity to an appropriate finite number of components. The DP mixture model also allows us to view our *infinite hidden relational model* as a generalization of nonparametric hierarchical Bayesian modeling to relational models (compare also, Xu et al., 2005). The combination of the hidden relational model and the DP mixture model is the infinite hidden relational model.

After presenting related work we will briefly introduce our preferred framework for describing relational models, i.e., the directed acyclic probabilistic entity relationship (DAPER) model. In Section 4 we will describe infinite hidden relational models and in Section 5 we introduce a modified Chinese restaurant sampling process to accommodate for the relational structure. In the subsequent sections we describe experimental results applying infinite hidden relational models to encode movie preference for users, to predict the functional class of a gene, and on a medical example, modeling the relationship between patients, diagnosis and procedures. In Section 9 we will present conclusions.

## 2 Related Work

Our approach can be related to some existing work. (Getoor et al., 2000) refined probabilistic relational models with class hierarchies, which specialized distinct probabilistic dependency for each subclass. (Rosen-Zvi et al., 2004) introduced an author-topic model for documents. The model implicitly explored the two relationships between documents and authors and document and words. (Kemp et al., 2004) showed a relational model with latent classes. (Carbonetto et al., 2005) introduced the nonparametric BLOG model, which specifies nonparametric probabilistic distributions over possible worlds defined by first-order logic. These models demonstrated good performance in certain applications. However, most are restricted to domains with simple relations. The proposed model goes beyond that by considering multiple related entities. In addition, the nonparametric nature allows the complexity of the model to be tuned by the model based on the available data set.

## 3 The DAPER Model

The DAPER model (Heckerman et al., 2004) formulates a probabilistic framework for an entity relationship database model. The DAPER model consists of entity classes, relationship classes, attribute classes and arc classes, as well as local distribution classes and constraint classes. Figure 1 shows an example of a DAPER model for a universe of students, courses and grades. The entity classes specify classes of objects in the real world, e.g. Student and Course shown as rectangles in Figure 1. The relationship class repre-sents interaction among entity classes. It is shown as a diamond-shaped node with dashed lines linked to the related entity classes. For example, the relationship, Take($s$, $c$) indicates that a student $s$ takes a class $c$. Attribute classes describe properties of entities or relationships. Attribute classes are connected to the corresponding entity/relationship class by a dashed line. For example, associated with courses is the attribute class Course.Difficulty. The attribute class $\theta$ in Figure 1 represents the parameters specifying the probability of student's grade in different configurations (i.e. course's difficulty and student's IQ). The arc classes shown as solid arrows from "parent" to "child" represent probabilistic dependencies among corresponding attributes. For example, the solid arrow from Student.IQ to Course.Grade specifies the fact that student's grade probabilistically depends on student's IQ. For more details please refer to (Heckerman et al., 2004). A relationship attribute might have the special attribute Exist with Exist= 0 indicating that the relationship does not exist (Getoor et al., 2003). Given particular instantiations of entities and relationships a ground Bayesian network can be formed which consists of all attributes in the domain linked by the resulting arc classes.

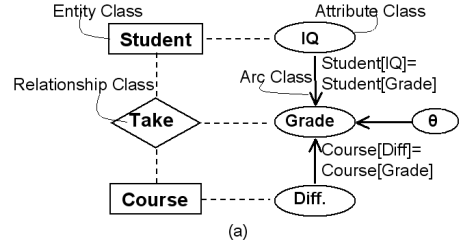

Figure 1: An example of DAPER model over university domain from Heckerman et al. (2004).

## 4 Infinite Hidden Relational Models

### 4.1 Hidden Relational Models

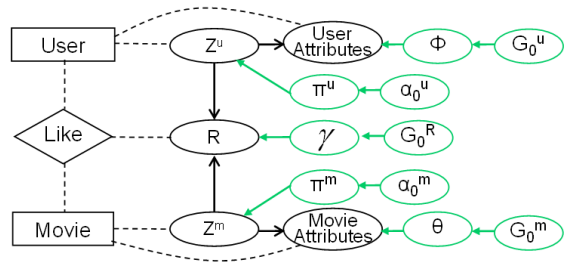

Figure 2: Infinite hidden relational Model on movie recommendation.

An example of an hidden relational model is shown in Figure 2. The example shows a movie recommendation system with entity classes User and Movie and relationship class Like. Furthermore, there are the attributes UserAttributes, MovieAttributes and R (rating) and various parameters and hyperparameters. The first innovation in our approach is to introduce for each entity a latent variable, in the example denoted as $Z^u$ and $Z^m$. They can be thought of as unknown attributes of the entities and are the parents of both the entity attributes and the relationship attributes. The underlying assumption is that if the latent variable was known, both entity attributes and relationship attributes can be well predicted. The most important result from introducing the latent variables is that now information can propagate through the ground network of interconnected latent variables. Let us consider the prediction of the relationship attribute $R$. If both the associated user and movie have strong known attributes, those will determine the state of the latent variables and the prediction for $R$ is mostly based on the entity attributes. In terms of a recommender system we would obtain a content-based recommendation system. Conversely, if the known attributes are weak, then the states of the latent variables for the user might be determined by the relationship attributes in relations to other movies and the states of those movies' latent variables. With the same argument, the states of the latent variables for the movie might be determined by the relationship attributes in relations to other users and the states of those users' latent variables. So by introducing the latent variables, information can globally distribute in the ground network defined by the relationship structure. This reduces the need for extensive structural learning, which is particularly difficult in relational models due to the huge number of potential parents. Note that a similar propagation of information can be observed in hidden Markov models in speech systems or in the hidden Markov random fields used in image analysis (Yedidia et a., 2005). In fact the hidden relational model can be viewed as a generalization of both for relational structures.

We now complete the model by introducing the parameters. First we consider the user parameters. Assume that $Z^u$ has $r^u$ states and that $\pi^u = (\pi_1^u, \ldots, \pi_{r^u}^u)$ are multinomial parameters with $P(Z^u = i) = \pi_i^u$ ($\pi_i^u \geq 0, \sum_i \pi_i^u = 1$). The multinomial parameters are drawn from a Dirichlet prior with $\pi^u \sim \mathrm{Dir}(\cdot | \alpha_0^u / r^u, \ldots, \alpha_0^u / r^u)$.

In the experiments all user attributes are assumed to be discrete and independent given $Z^u$. Thus, a particular user attribute $A^u$ with $r$ states is a sample from a multinomial distribution with $P(A^u = i) = \phi_i$ and

$$(\phi_1, \ldots, \phi_r) \sim G_0^u = \mathrm{Dir}(\cdot | \beta_1^*, \ldots, \beta_r^*).$$

It is also convenient to re-parameterize

$$\beta_0 = \sum_{k=1}^r \beta_k^* \quad \beta_k = \frac{\beta_k^*}{\beta_0} \quad k = 1, \ldots, r$$

and $\beta = \{\beta_1, \ldots, \beta_r\}$. In the application, we assume a neutral prior with $\beta_k = 1/r$, which represents our prior belief in the fact that the multinomial parameters should be equal. $\beta_0$ is a parameter indicating how strongly we believe that the prior distribution represented by $\beta$ should be true. We finely tune $\beta_0$ using a cross validation procedure.

The parameters for the entity class Movie and the relationship class Like are defined in an equivalent way. Note, that for the relationship attribute $R$, $r^u \times r^m$ parameter vectors $\gamma$ are generated.

## 4.2 Infinite Hidden Relational Models

The latent variables play a key role in our model and in many applications, we would expect that the model might require a large number of states for the latent variables. Consider again the movie recommendation system. With little information about past ratings all users might look the same (movies are globally liked or disliked), with more information available, one might discover certain clusters in the users (action movie aficionados, comedy aficionados, ...) but with an increasing number of past ratings the clusters might show increasingly detailed structure ultimately indicating that everyone is an individual. It thus makes sense to permit an arbitrary number of latent states by using a Dirichlet process mixture model. This permits the model to decide itself about the optimal number of states for the latent variables. In addition, the infinite hidden relational model can now also be viewed as a direct generalization of a nonparametric hierarchical Bayesian approach. For an introduction to Dirichlet processes see for example Teh et al. (2004) and Tresp (2006). For our discussion is suffices to say that we obtain an infinite hidden relational model by simply letting the number of states approach infinity, $r^u \to \infty$, $r^m \to \infty$. Although a model with infinite numbers of states and parameters cannot be represented, it turns out that sampling in such model is elegant and simple, as shown in the next section.

In the Dirichlet mixture model, $\alpha_0$ determines the tendency of the model to either use a large number or a small number of states in the latent variables, which is also apparent from the sampling procedures described below. In our experiments, we found that $\alpha_0$ is rather uncritical and was fixed for all models to be equal to ten.

# 5 Sampling in the Infinite Hidden Relational Model

Although a Dirichlet process mixture model contains an infinite number of parameters and states, the sampling procedure only deals with a growing but finite representation. This sampling procedure is based on the Chinese restaurant process (CRP) where a state of a latent variable is identified as a cluster, i.e., a table in a restaurant. We will now describe how the CRP is applied to the infinite hidden relational model. The procedure differs from the standard model by the sampling of the *relational attribute* where two CRP processes are coupled. We omit the description of the sampling of the attributes which is straightforward, given parameter samples.

- The first user is assigned to the user cluster 1, $Z_1^u = 1$; an associated parameter is generated $\phi_1 \sim G_0^u$. Similarly, the first movie is assigned to the movie cluster 1, $Z_1^m = 1$; an associated parameter is generated $\theta_1 \sim G_0^m$.

- The parameter $\gamma_{1,1} \sim G_0^R$ is drawn for the attributes of the relation $R$ between the first user cluster and the first movie cluster.

- With probability $1/(1 + \alpha_0^u)$, the second user is also assigned at the user cluster 1, $Z_2^u = 1$, and inherits $\phi_1$ and $\gamma_{1,1}$; with probability $\alpha_0^u/(1 + \alpha_0^u)$ the user is assigned at cluster 2, $Z_2^u = 2$, and new parameters are generated $\phi_2 \sim G_0^u$ and $\gamma_{2,1} \sim G_0^R$.

- Equivalently, the second movie is generated. With probability $1/(1+\alpha_0^m)$, the second movie is assigned at the movie cluster 1, $Z_2^m = 1$, and inherits $\theta_1$ and $\gamma_{1,1}$; with probability $\alpha_0^m/(1+\alpha_0^m)$ the movie is assigned at cluster 2, $Z_2^m = 2$, and new parameters are generated $\theta_2 \sim G_0^m$, $\gamma_{1,2} \sim G_0^R$ and $\gamma_{2,2} \sim G_0^R$ (if the second user cluster has appeared so far).

- We continue this process, after $N^u$ users and $N^m$ movies have been generated, $K^u$ user clusters and $K^m$ movie clusters appear, $N_i^u$ users are assigned to the user cluster $i$, $N_j^m$ movies are assigned to the movie cluster $j$.

- New user:
  - The user $N^u + 1$ is assigned with probability $\frac{N_i^u}{N^u + \alpha_0^u}$ to a previously existing cluster $i$ and inherits $\phi_i$ and $\{\gamma_{i,l}\}_{l=1}^{K^m}$. Thus: $Z_{N^u+1}^u = i$, $N_i^u \leftarrow N_i^u + 1$;
  - With probability $\frac{\alpha_0^u}{N^u + \alpha_0^u}$ the user is assigned to a new cluster $K^u + 1$. Thus: $Z_{N^u+1}^u = K^u + 1$, $N_{K^u+1}^u = 1$.
  - For the new user cluster, new parameters are generated: $\phi_{K^u+1} \sim G_0^u$ and $\gamma_{K^u+1,l} \sim G_0^R$, $l = 1 : K^m$. $K^u \leftarrow K^u + 1$.

- New movie:
  - The generative process for the movie $N^m + 1$ is equivalent.
  - The movie $N^m + 1$ is assigned with probability $\frac{N_j^m}{N^m + \alpha_0^m}$ to a previously existing cluster $j$ and inherits $\theta_j$ and $\{\gamma_{l,j}\}_{l=1}^{K^u}$. Thus: $Z_{N^m+1}^m = j$, $N_j^m \leftarrow N_j^m + 1$;
  - With probability $\frac{\alpha_0^m}{N^m + \alpha_0^m}$ the movie is assigned to a new cluster $K^m + 1$. Thus: $Z_{N^m+1}^m = K^m + 1$, $N_{K^m+1}^m = 1$.
  - For the new movie cluster, new parameters are generated: $\theta_{K^m+1} \sim G_0^m$ and $\gamma_{l,K^m+1} \sim G_0^R$, $l = 1 : K^u$. $K^m \leftarrow K^m + 1$.

The previous procedure generates samples from the generative model. Now we consider sampling from a model given data, i.e. given a set of movie attributes, user attributes and ratings. We assume that the model has $U$ users and $M$ movies and that some instances of $A^u$, $A^m$, $R$ are known. The goal is now to generate samples of the parameters $\phi, \theta, \gamma$, the latent variables $Z^u$ and $Z^m$, which allows us to then make predictions about unknown attributes. We exploit Gibbs sampling inference based on the Chinese restaurant procedure as described in the Appendix. Note, that since the attributes appear as children, unknown attributes can be marginalized out and thus removed from the model, greatly reducing the complexity. Although the DP mixture model contains an infinite number of states, in the Gibbs sampling procedure only a finite number of states is ever occupied, providing an estimate of the true underlying number of clusters (Tresp, 2006). Details on the Gibbs sampler can be found in the Appendix.

# 6 Experiment on MovieLens

We first evaluate our model on the MovieLens data which contains movie ratings from a large number of users (Sarwar et al. 2000). The task is to predict whether a user likes a movie. There are two entity classes (User and Movie) and one relationship class (Like: users like movies). The User class has several attribute classes such as Age, Gender, Occupation. The Movie class has attribute classes such as Published-year, Genres and so on. The relationship has an additional attribute $R$ with two states: $R = 1$ indicates that the user likes the movie and $R = 0$ indicates otherwise. The model is shown as Figure 2. In the data set, there are totally 943 users and 1680 movies. In addition, user ratings on movies are originally recorded on a five-point scale, ranging from 1 to 5. We transfer the ratings to be binary, *yes* if a rating higher than the average rating of the user, and vice versa. Model performance is evaluated using prediction accuracy. The experimental results are shown in Table 1. First we did experiments ignoring the attributes of the users and the items. We achieved an accuracy of 69.97% (E3). This is significantly better in comparison to approaches using one-sided collaborative filtering by generalizing across users (E1) leading to an accuracy of 64.22% or by generalizing across items (E2) leading to an accuracy of 64.66%. When we added information about the attributes of the users and the model, the prediction accuracy only improved insignificantly to 70.3% (E4): the reason is that the attributes are weak predictors of preferences as indicated by the bad performance of the SVM prediction (54.85% accuracy, E5) which is solely based on the attributes of the users and the items.

Table 1: The accuracy of predicting relationships between Users and Movies

| Method | Accuracy(%) |
|---|---|
| E1: Collaborative filtering 1 | 64.22 |
| E2: Collaborative filtering 2 | 64.66 |
| E3: Infinite hidden relational model without attributes | 69.97 |
| E4: Infinite hidden relational model | 70.3 |
| E5: Content based SVM | 54.85 |

## 7  Experiment on Medical Data

The second experiment is concerned with a medical domain. The proposed model is shown in Figure 3(a). The domain includes three entity classes (Patient, Diagnosis and Procedure) and two relationship classes (Make: physician is making a diagnosis and Take: patient taking a procedure). A patient typically has both multiple procedures and multiple diagnoses. The Patient class has several attribute classes including Age, Gender, PrimaryComplaint. To reduce the complexity of Figure 3(a), patient characteristics are grouped together as PatientAttributes (these attributes are not aggregated in learning and inference). The DiagnosisAttributes contain the class of the diagnosis as specified in the ICD-9 code and the ProcedureAttributes contain the class of the procedure as specified in the CPT4 code. Both the relationships between the patients and the procedures and the relationships between the patients and the diagnoses are modeled as existence uncertainty. $R^{pa,pr} = 1$ means that the patient received the procedure and $R^{pa,pr} = 0$ indicates otherwise. Equivalently, $R^{pa,dg} = 1$ means that the patient received the diagnosis and $R^{pa,dg} = 0$ indicates otherwise. In the data, there are totally 14062 patients, 703 diagnoses and 367 procedures. The infinite hidden relational model contains three DPs, one for each entity class. We compare our approach with two models. The first one is a relational model using reference uncertainty (Getoor et al., 2003) without a latent variable structure. The second comparison model is a content based Bayesian network. In this model, only the attributes of patients and procedures determine if a procedure is prescribed.

We test model performances by predicting the application of procedures. ROC curve is used as evaluation criteria. In the experiment we selected the top $N$ procedures recommended by the various models. Sensitivity indicates how many percent of the actually being performed procedures were correctly proposed by the model. (1-specificity) indicates how many of the procedures that were not actually performed were recommended by the model. Along the curves, the $N$ was varied from left to right as $N = 5, 10, \ldots, 50$.

In the experiment we predict a relation between a patient and a procedure *given her first procedure*. The corresponding ROC curves (averaged over all patients) for the experiments are shown in Figure 4. The infinite hidden relational model (E3) exploiting all relational information and all attributes gave best performance. When we remove the attributes of the entities, the performance degrades (E2). If, in addition, we only consider the one-sided collaborative effect, the performance is even worse (E1). (E5) is the pure content-based approach using the Bayesian network. The results show that entity attributes are a reasonable predictor but that the performance of the full model cannot be achieved. (E4) shows the results of relational model using relational uncertainty, which gave good results but did not achieve the performance of the infinite hidden relational model. Figures 5 shows the corresponding plots for a selected class of patients; patients with prime complaint *respiratory problem*. The results exhibit similar trends.

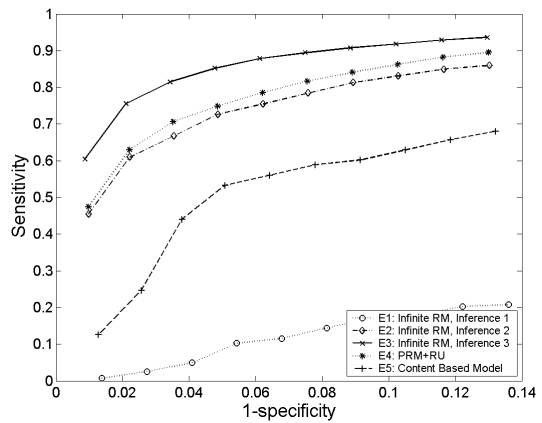

Figure 4: ROC curves for predicting procedures.

## 8  Experiment on Gene Data

The third evaluation is performed on the yeast genome data set of KDD Cup 2001 (Cheng et al. 2002). The goal is to evaluate the proposed algorithm to model the dependencies between relationships instantiated from various relationship classes. The genomes in several organisms have been sequenced. Genes code for proteins, the proteins localize in various parts of *the cells* and interact with one another to perform crucial functions. Traditionally, the functions of genes/proteins are predicted by comparing with characterized genes/proteins in sequence similarity. But only 52% of 6449 yeast proteins have been characterized. Of the remaining, only 4% show strong similarity with the known ones at the sequence level. It is therefore necessary to integrate other information to characterize genes/proteins. The goal of our experiment is to predict functions of genes based on the information not only at the gene-level but also at the protein-level. The data set consists of two relational tables that are produced from the original seven relational tables. One table speci-fies a variety of properties of genes or proteins. These properties include *chromosome, essential, phenotype, motif, class, complex and function*. *Chromosome* expresses the chromosome on which the gene appears. *Essential* specifies whether organisms with a mutation in this gene can survive. *Phenotype* represents the observed characteristics of organisms with differences in this gene. *Class* means the structural category of the protein for which this gene codes. *Motif* expresses the information about the amino acid sequence of the protein. The value of property *complex* specifies how the expression of the gene can complex with others to form a larger protein. The other table contains the information about interactions between genes. A gene typically has multiple complexes, phenotypes, classes, motifs and functions, respectively but only one property essential and one property chromosome. An example gene is shown in Table 2. To keep the multi-relational nature of the data, we restore the original data structure. There are six entity classes (Gene, Complex, Phenotype, Class, Motif and Function) and six relationship classes (Interact: genes interact with each other, Have: genes have functions, Observe: phenotype are observed for the genes, Form: which kinds of complex is formed for the genes, Belong: genes belong to structural classes, Contain: genes contain characteristic motifs). Gene class has attribute classes such as Essential, Chromosome, etc. The attributes of other entity classes are not available in the data set. A hidden attribute is added into each entity class. All relationships are modeled as existence uncertainty. Thus each relationship class has additional attribute $R$ with two states. The state of $R$ indicates whether the relationship exists or not. The task of function prediction of genes is therefore transformed to the relationship prediction between genes and func-tions. The data set totally contains 1243 genes. A subset (381 genes) is withheld for testing in the KDD Cup 2001. The remaining 862 genes are provided to participants. In the data, there are 56 complexes, 11 phenotypes, 351 motifs, 24 classes and 14 functions. There are two main challenges in the gene data set. First, there are many types of relationships. Second, there are large numbers of objects, but only a small number of known relationships.

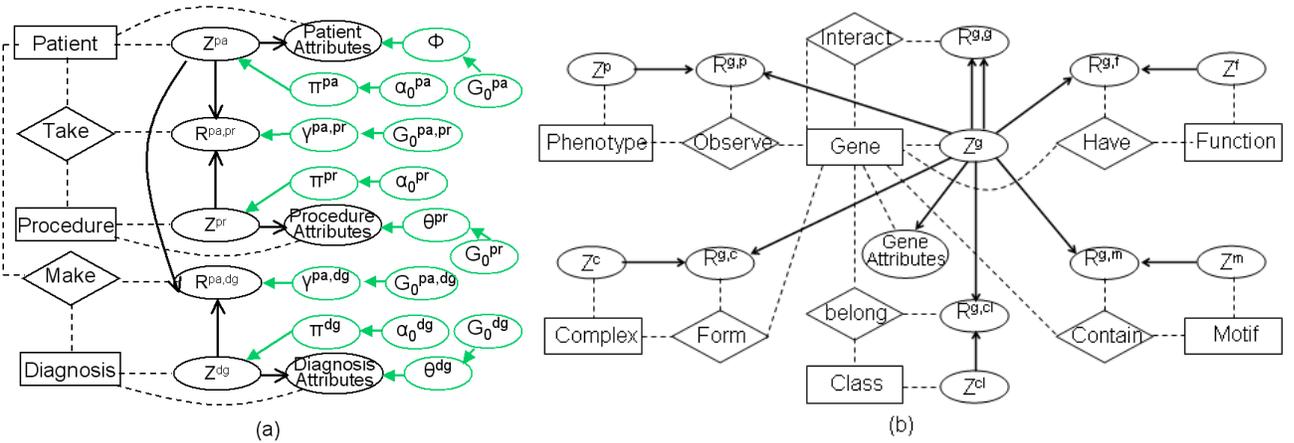

Figure 3: Infinite hidden relational model for (a) a medical database and (b) a gene database.

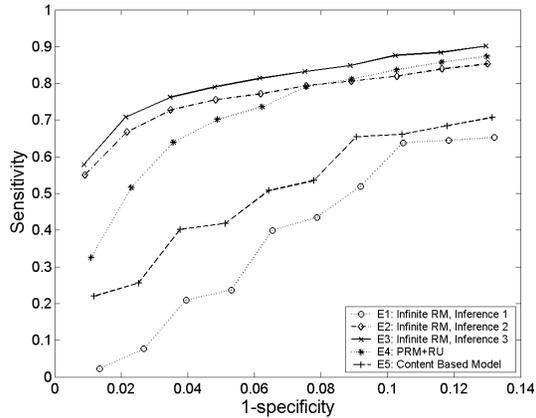

Figure 5: ROC curves for predicting procedures on a subset of patients with prime complaint *respiratory problem*.

Table 2: An example gene

| Attribute | Value |
|---|---|
| Gene ID | G234070 |
| Essential | Non-Essential |
| Class | 1, ATPases 2, Motorproteins |
| Complex | Cytoskeleton |
| Phenotype | Mating and sporulation defects |
| Motif | PS00017 |
| Chromosome | 1 |
| Function | 1, Cell growth, cell division and DNA synthesis 2, Cellular organization 3, Cellular transport and transprotmechanisms |
| Localization | Cytoskeleton |

The proposed model applied to the gene data is shown in Figure 3(b). The existence of any relationship depends on the hidden states of the corresponding entities. The information about a variety of relationships of Gene is propagated via the hidden attribute of Gene. The model is optimized using 862 genes, and is applied on the testing data. The experiment results are shown in Table 3. There were 41 groups that participated in the KDD Cup 2001 contest. The algorithms include naive Bayes, k-nearest neighbor, decision tree, neural network, SVM, and Bayesian networks, etc. and technologies such as feature selection, boosting, cross validation, etc., were employed. The performance of our model is comparable to the best results. The winning algorithm is a relational model based on inductive logic programming. The infinite hidden relational model is only slightly worse (probably not significantly) if compared to the winning algorithm.

Table 3: Prediction of gene functions

| Model | Accuracy (%) | True Positive Rate (%) |
|---|---|---|
| Infinite model | 93.18 | 72.8 |
| Kdd cup winer | 93.63 | 71.0 |

In the second set of experiments, we investigated the influence of a variety of relationships on the prediction of functions. We perform the experiments by ignoring a specific kind of known relationships. The result is shown in Table 4. When a specific type of known relationship is ignored, lower accuracy indicates higher importance of this type of relationship. One observation is that the most important relationship is *Complex*, specifying how genes complex with another genes to form larger proteins. The second one is the interaction relationships between genes. This coincide with the lesson learned from KDD Cup 2001 that protein interaction information is less important in function prediction. This lesson is somewhat surprising since there is a general belief in biology that the knowledge about regulatory pathways is helpful to determine the functions of genes.

Table 4: The importance of a variety of relationships in function prediction of genes

| Ignored relationships | Accuracy(%) | Importance |
|---|---|---|
| Complex | 91.13 | 197 |
| Interaction | 92.14 | 100 |
| Class | 92.61 | 55 |
| Phenotype | 92.71 | 45 |
| Attributes of gene | 93.08 | 10 |
| Motif | 93.12 | 6 |

## 9 Conclusions and Extensions

We have introduced the infinite hidden relational model. The model showed encouraging results on a number of data sets. We hope that infinite hidden relational model will be a useful addition to relational modeling by allowing for flexible inference in a relational network reducing the need for extensive structural model search. Implicitly, we have assumed a particular sampling scheme, i.e., that entities are independently sampled out of unspecified populations. In this context our model permits generalization but it might fail if this assumption is not reasonable or if the sampling procedure changes in the test set. We have focussed on an explicit modeling of the relation between *pairs* of entities but our model can easily be generalized if more than two entities are involved in a relation. As part of our future work we will explore and compare different approximate inference algorithms.

## Appendix: Inference based on Gibbs Sampling

We assume that users are assigned to the first $K^u$ states of $Z^u$ and movies are assigned to the first $K^m$ states of $Z^m$. We can do this without loss of generality by exploiting exchangeability. Note, that $K^u \leq U$ and $K^m \leq M$. If during sampling a state becomes unoccupied that state is removed from the model and indices are re-assigned. To simplify the description of sampling we will assume that this does not occur and that currently no state is occupied by exactly one item (just to simplify book keeping).

Gibbs sampling updates the assignment of users and movies to the states of the latent variable and re-samples the parameters. In detail:

1. Pick a random user $j$. Assume that for $N_k^u$ users, $Z^u = k$ without counting user $j$.

    (a) Then, we assign state $Z_j^u = k$ with $N_k^u > 0$ with probability proportional to
    $$P(Z_j^u = k | \{Z_l^u\}_{l \neq j}^U, D_j^u, \phi, \gamma, Z^m) \propto$$
    $$N_k^u P(D_j^u | \phi_k, \gamma_{k,*}, Z^m)$$

    (b) Instead, a new state $K^u + 1$ is generated with probability proportional to
    $$P(Z_j^u = K^u + 1 | \{Z_l^u\}_{l \neq j}^U, D_j^u, \phi, \gamma, Z^m) \propto$$
    $$\alpha_0^u P(D_j^u)$$

    (c) In the first case, the j-th model inherits the parameters assigned to state $k$: $\phi_k, \gamma_{k,1}, \ldots \gamma_{k,K^m}$

    (d) In the latter case: new parameters are generated following
    $$P(\phi_{K^u+1} | D_j^u)$$
    and
    $$P(\gamma_{K^u+1,l} | D_j^u, Z^m), l = 1, \ldots, M$$

2. Pick a random movie $i$. Updates the latent variables of $Z_i^m$. The sampling is equivalent to the sampling of $Z^u$, above.

3. Occasionally (typically less often than the updates for the latent variables): Update the parameters, $\phi, \theta, \gamma$ from posterior distribution based on all the data assigned to the corresponding a state, resp. pairs of states.

In the algorithm, we used the following definitions (terms involving entity attributes or relationship attributes which are not known drop out of the equations)

$$P(D_j^u | \phi_k, \gamma, Z^m) = P(A_j^u | \phi_k) \prod_{l=1}^M P(R_{j,l} | \gamma_{k, Z_l^m})$$
$$P(\phi | D_j^u) \propto P(A_j^u | \phi) \, G_0^u(\phi)$$
$$P(\gamma | D_j^u) \propto \prod_{l=1}^M P(R_{j,l} | \gamma) \, G_0^R(\gamma)$$

The algorithm easily generalizes to multiple relations as described in Section 7 and Section 8. $A_j^u$ denotes all known attributes of user $j$. Definitions for the movies are equivalent.

The most expensive term in the algorithm is in step 1 (a) which scales proportional to the number of known entity and relational attributes of the involved entity and is proportional to the number of occupied states.

We are currently exploring various sampling schemes and deterministic approximations. An extensive comparison will be available on the web shortly. The results reported in this paper were obtained by using the deterministic approximation described in Tresp and Yu, 2004.